\documentclass[runningheads]{llncs}
\usepackage{graphicx}
\usepackage{natbib}
\usepackage{tkz-tab}
\usepackage{multicol}
\usepackage[english]{babel}
\usepackage[colorlinks=true, allcolors=blue]{hyperref}
\usepackage{tikz}
\usepackage{float}
\usepackage{caption}
\usepackage{graphicx} 
\usepackage{amssymb} 
\usepackage{hyperref}
\usepackage{enumerate}
\usepackage{neuralnetwork}
\usepackage{cancel}
\usepackage{multirow}
\usepackage{float}
\usepackage{pgfplots}
\usepackage{bm}
\usepackage{multicol}
\usepackage{soul}
\usepackage{array}
\usepackage{everysel}
\usepackage{makecell} 
\usepackage{multirow}
\usepackage{ragged2e}
\newcommand{\comment}[1]{}
\usepackage{comment}

\usepackage{lscape}
\usepackage{rotating}

\begin{document}

\title{Surrogate Neural Networks Local Stability for Aircraft Predictive Maintenance}
\titlerunning{NN Local Stability for Aircraft Predictive Maintenance}

\author{
    Mélanie Ducoffe\inst{1} \and
    Guillaume Povéda\inst{1} \and
    Audrey Galametz\inst{2} \and
    Ryma Boumazouza\inst{1} \and
    Marion-Cécile Martin\inst{2} \and 
    Julien Baris\inst{1} \and
    Derk Daverschot\inst{2} \and
    Eugene O'Higgins\inst{2}
}
\institute{
Airbus Operations SAS, France 
\and Airbus Operations GmbH, Germany}
\authorrunning{M. Ducoffe, G. Povéda et al.}

\maketitle

\begin{abstract}
Surrogate Neural Networks are nowadays routinely used in industry as substitutes for computationally demanding engineering simulations (e.g.,~in structural analysis). They allow to generate faster predictions and thus analyses in industrial applications e.g.,~during a product design, testing or monitoring phases. Due to their performance and time-efficiency, these surrogate models are now being developed for use in safety-critical applications. Neural network verification and in particular the assessment of their robustness (e.g.,~to perturbations) is the next critical step to allow their inclusion in real-life applications and certification. We assess the applicability and scalability of empirical and formal methods in the context of aircraft predictive maintenance for surrogate neural networks designed to predict the stress sustained by an aircraft part from external loads. The case study covers a high-dimensional input and output space and the verification process thus accommodates multi-objective constraints. We explore the complementarity of verification methods in assessing the local stability property of such surrogate models to input noise. We showcase the effectiveness of sequentially combining methods in one verification `pipeline' and demonstrating the subsequent gain in runtime required to assess the targeted property.


 

\end{abstract}

\section{Introduction}

There have been significant advances in the development of verification methods to assess the robustness of neural networks (NN). Formal methods are, in particular, considered as key approaches to be explored and matured (see \S\ref{related_work}). They are expected to provide model guarantees that could allow industries, such as aviation, to meet the emerging certification requirements for the use of NN, in particular for safety-critical systems. Notable references, including guidelines from certification and aviation standardization entities, are advocating for the use of formal methods as a mean of compliance during the learning process management and inference model verification phases (see, e.g.~the European Aviation Safety Agency (EASA) recent concept paper `First usable guidance for Level 1 machine learning applications\footnote{https://www.easa.europa.eu/en/easa-concept-paper-first-usable-guidance-level-1-machine-learning-applications-proposed-issue-01pdf \label{EASA}}). A concrete example of this use is introduced in the ForMuLA report (Formal Methods use for Learning Assurance), published from the partnership between EASA and Collins Aerospace \cite{easa2023c}. It includes an application of formal methods to an industrial use case of prediction of the remaining life of aeronautical components and illustrates how these  techniques can be used to assess the safety of machine-learning (ML) based systems.  

The focus of published works on the applications of formal methods to ML models has predominantly been on classification tasks. There has been comparatively less emphasis on regression tasks, especially those tailored for industrial applications. It is however crucial to extend the implementation of these methods to regression challenges in regard to the increasing use of neural network surrogate models in industry, in particular in aviation where many are developed to act as substitute to computationally-demanding simulations (see \S\ref{usecase}).

In this paper, we are conducting the robustness assessment of NN surrogate models designed in the context of civil aviation and predictive aircraft maintenance. The case study presents unique specificities (e.g.~regression task, multiple inputs and outputs etc.), rarely explored in the current literature. In this case study, we explore the adaptability and scalability of several families of verification techniques to the use case at hand. We further build upon their strengths and limitations to optimise their sequential and complementary use in order to perform the complete NN robustness assessment on a representative test set.

Section~\ref{related_work} summarizes the recent advances in NN verification. Section~\ref{usecase} introduces the case study of aircraft loads-to-stress prediction along with the tested models and stability property to be ensured. Section~\ref{methodcomb} describes the combination via the sequential use of techniques we employ in order to perform the NN stability assessment. Section~\ref{sec:experiments} details the experiments and illustrates the verification process on a few NN models. Sections~\ref{sec:results} and \ref{sec:conclusions} present findings and conclusions.

\section{Related Work}
\label{related_work}

NN are brittle. The emergence of the field of adversarial machine-learning has shown that models can be easily fooled, even by small perturbations to input data \cite{goodfellow2014explaining}. A series of work has subsequently attempted to craft defense mechanisms against adversarial attacks e.g.,~\cite{madry2017towards}. Ultimately, however, these techniques do not provide guarantee that a neural network is robust to perturbations but only that counterexamples to a robustness requirement to be guaranteed can be generated. 

The introduction of adversarial attacks has motivated the ramp-up of the applicability of formal methods to AI models, in particular to assess the robustness of neural networks against small `local' perturbations, often represented by $l_p$ norm balls. Two families of verification methods have since been matured: {\it (i)} incomplete techniques e.g., \cite{katz2017reluplex, katz2019marabou, kouvaros2021towards} that can prove the presence/absence of counterexamples to a given property. They are usually computationally expensive and scale poorly with large networks. {\it (ii)} complete techniques, e.g.,  \cite{NEURIPS2018_f2f44698, DBLP:journals/pacmpl/SinghGPV19, zhang2018efficient, xu2021fast}, that provide approximated bounds to model prediction. They commonly perform faster and thus scale better with larger models. They can however severely over approximate bound estimates of model output, leading to their lack of convergence for a given robustness status. 

Verification tools are now commonly combining complete and incomplete techniques \cite{wu2024marabou} in order to speed the verification process. However, to the best of our knowledge, there has been very limited published works illustrating and quantifying the impact (e.g. on computing time) this combination brings. As model size and input dimensionality keep increasing, scalability and verification optimisation will however be of essence for the adoption of formal methods in the verification, validation and certification processes of AI-based industrial systems.    





\section{Case study: Aircraft Loads-to-Stress Prediction}
\label{usecase}

\subsection{Description}

In aeronautics, numerical simulations are regularly used to model complex physical phenomena in systems or structures (using, e.g.~a finite element discretization approach) and understand and anticipate their behaviour. The computational cost of such simulations has however prevented their use in real-time, including their embeddability in products or on-line processes. Their application therefore remains limited e.g.~to a system design phase. The use of such simulation could however greatly improve productivity and operational cost, e.g.~in the domain of aircraft predictive maintenance. 

Structural maintenance programs (e.g.~replacement of parts etc.) are currently based on conservative assumptions: they often assume aircrafts of a fleet have sustained a similar flight history and, thus, operational fatigue, and adopt worse case scenario, often overly preventive, maintenance actions for the whole fleet. One can therefore see the benefit in introducing a more optimised maintenance program, tailored to individual aircraft in order to reduce individual maintenance cost while ensuring that the safety of operations is maintained at the highest level. Such custom solutions, however, require the analysis of large amount of operational data from flight past history, the complete modelisation of the sustained aircraft fatigue and the prediction of the impact it had on its different parts in order to anticipate the required maintenance operations. These simulations involve computationally-demanding physics and mathematical modeling that may be unfeasible to use in real-life operational settings.

NN are a game changer. More and more studies are showing the value of NN-based surrogates to approximate numerical simulators \cite{sudakov2019artificial}. Their more systematic use in industry however requires the maturation of new verification processes, specific to machine-learning components. The current lack of consensus on the tools to conduct such verification dramatically limits their applications, especially as components or enablers of safety critical applications. In this analysis, we investigate an example of NN surrogate models trained to predict the level of stress in different parts of an aircraft structure from sustained external loads. More details on the model function is provided in \cite{O’Higgins2020}. 
The accurate evaluation of the stresses is a key enabler for maintenance optimization. 

\vspace{-4mm}
\subsection{Tested models}

We evaluate NNs composed of two hidden layers ($h$) with $165$ neurons ($r$) and a dense output layer of size $81$. Each hidden layer is followed by \textit{ReLU} activation functions. The NNs predict $81$ normalised stress outputs ($k$) from $216$ normalised loads input variables ($n$). In order to investigate the potential impact of training epochs on model robustness, we train a series of models with increasing number of epochs ($5$ models from $2$ to $10000$ epochs). These feed-forward NNs are functions $f : \mathbb{R}^n \mapsto \mathbb{R}^k$. Let $x$ be an input vector of dimension $n$ and $W$  and $b$ the weights and biases of the network respectively. $f$ is the composition of linear functions denoted $l_i: x\mapsto W_i\cdot x + b_i$, followed by element-wise non linear activation functions denoted $\sigma$. We only consider \textit{ReLU} activations, such that $\sigma: x \in \mathbb{R}^n\mapsto [max(x_j, 0)]_{j=1}^{r}$. 

%

The NN tested in the present study are research prototypes and earlier, less mature versions of the models currently pushed at Airbus. The models that will be used in predictive maintenance settings are currently being optimised, in part supported by robustness analysis such as the one presented here. 

\vspace{-4mm}
\subsection{Property to be ensured: Local stability}
\label{sec:property}

A NN is locally stable if its predictions in the immediate vicinity of a test data point are consistent i.e.,~within a small value range allowed by the safety constrains. In engineering-driven domains such as aeronautics, the need for local stability of NN prediction is obvious and thus, a critical requirement. Model designers indeed want to ensure that a model reflects the continuous characteristics of physics-driven systems and phenomena. In the present case study, equivalent load values should cause equivalent stress intensities sustained by an aircraft part. The assessment of local stability is one of the robustness properties EASA lists as part of a model learning process verification phase (see \ref{EASA}).

Local stability (and thus safety) requirements for the case study at hand are enunciated in property~\ref{property:stability}. In this particular case study, they are essentially split in two zones. We will refer to this property as the `bow tie' owing to its distinctive shape. The belonging to a given zone (`Knot' or `Wing') for a data point $x$ and a given output index $i$ is defined by the output value $f_i(x)$. \\
\vspace{-5mm}
\begin{property}
\label{property:stability}Local stability property a.k.a.,~the `bow tie'

\noindent Consider:
\begin{itemize}
    \item a neural network $f$,
    \item an input sample $x\in \mathbb{R}^n$,
    \item i, the $i^{th}$ output of the network $f_i(x)$ where $i \in [1, k]$,
    \item $p_{inp}$, a local input perturbation and  
    \item $x'$, a perturbed input such that $x' \in \big[ x - p_{inp}.|x|, x + p_{inp}.|x|\big]$\\
\end{itemize}

\begin{figure}[h!]
\begin{center}
    \includegraphics[width=.65\textwidth]{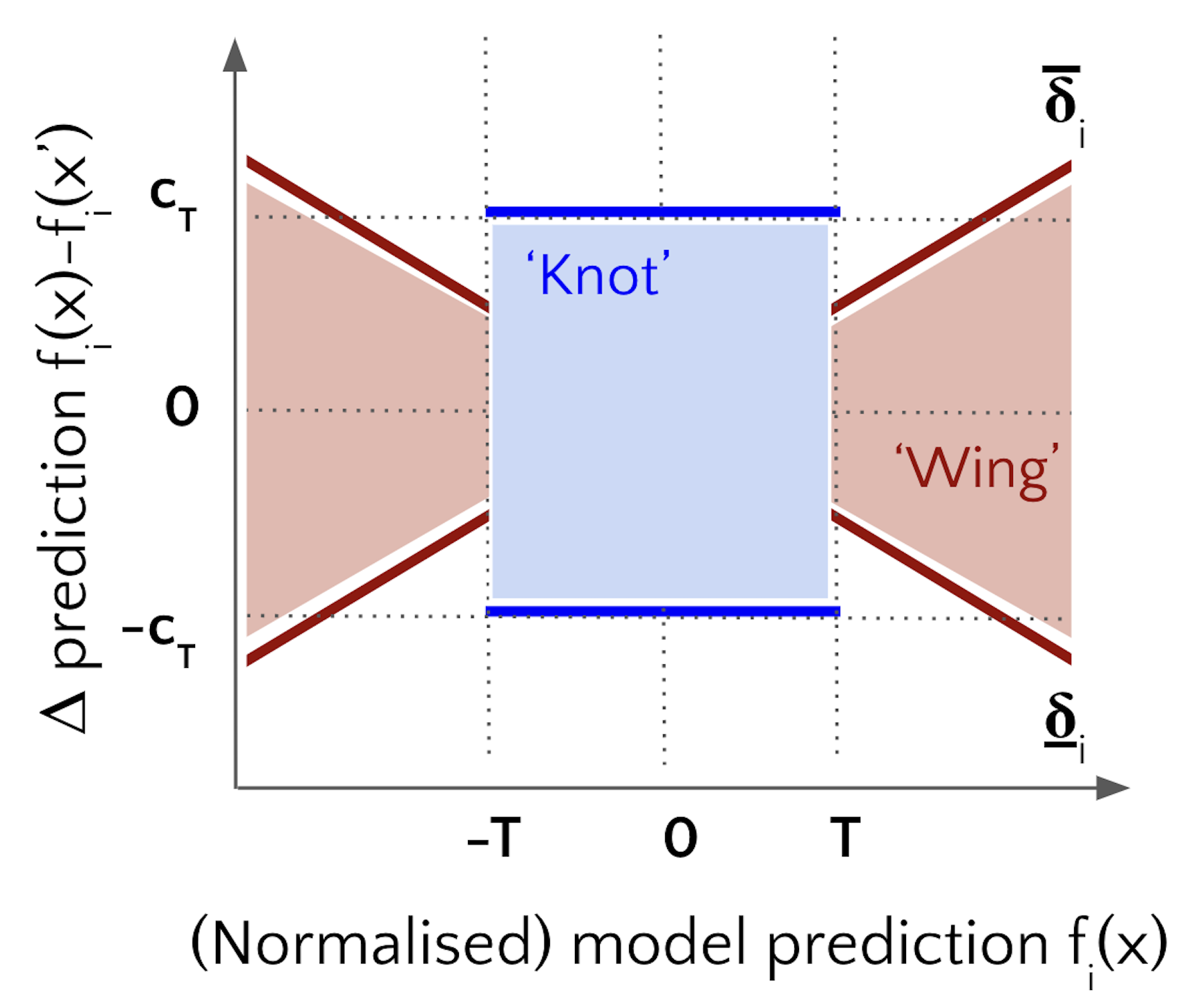}
\end{center}
\vspace{-10mm}
\end{figure}
    
\noindent The NN local stability is ensured if, for any $x'$: 
\begin{align*}
    \forall i\in [1, k], \ 
    &\underline{\delta_i} \leq f_i(x')-f_i(x) \leq \overline{\delta_i} \\
    &\text{with} \ [\underline{\delta_i}, \overline{\delta_i}]= \begin{cases}
            [- c_T , c_T] & \text{if } |f_i(x)| \leq T\\
            & \text{(`Knot')}\\
            [-p_{out}.|f_i(x)|,p_{out}.|f_i(x)|]  &\text{else} \\
            & \text{(`Wings')}\\
        \end{cases}
    \end{align*}

\noindent where 

\begin{itemize}
    \item $\underline{\delta_i}$ and $\overline{\delta_i}$ are respectively upper and lower constrains on deviation from original prediction,
    \item $T$ is a threshold defining two zones of the output space with distinct requirements on stability i.e.,~$|f_i(x)| \leq T$ \text{(`Knot')} and $|f_i(x)| > T$ \text{(`Wings')},
    \item $c_T$ is the set constraint on $\delta_i$ in the knot and
    \item $p_{out}$ are constrains on output perturbation in the wings.
\end{itemize}
\vspace{-2mm}

\end{property}

We adopt $T = 10$ ($T < 10$ is considered as the small loads regime), $c_T = 1$ and $p_{inp} = p_{out} = 0.05$, taking into account safety requirements derived from the domain expertise of aircraft structure experts involved in the development of the maintenance solution. In the present work, we adopt constrains that are index-independent i.e.,~$T$, $c_T$, $p_{inp}$ and $p_{out}$ have the same value for all output indexes. We also adopt a direct match between input perturbations and output constrains ($p_{inp} = p_{out}$). The implemented solution is however highly parameterised and the verification procedure can easily be adapted as the stability property is refined and optimised by domain experts.



\section{Method combination for local stability assessment}
\label{methodcomb}



Before the advent of formal NN verification methods, the model local stability was assessed at Airbus via a random sampling of its immediate `neighborhood' and corresponding predictions. This `brute force' approach is both partial and computational expensive due to the necessity to homogeneously cover via sampling a high number of perturbed input variables. There was therefore a crucial need for formal, sound and time-efficient means to assess the NN stability. 

Formal verification involves both complete and incomplete methods. Incomplete methods can either determine the property is verified for a given test point or are unable to conclude about its robustness i.e.,~it can not provide definite guarantees that a property is violated. Complete methods provide guarantees that a property is either verified or not. They are only inconclusive when the verification time required to reach a conclusion exceeds the time limit (`timeout') set for the verification. Both are `sound' techniques meaning that they can never provide an incorrect answer. In a complementary manner, empirical techniques such as adversarial attacks exploit model vulnerabilities and find small input perturbations that will make a model produce incorrect and/or unsafe outputs. 

Exploring a combination of techniques and using their strengths and limitations to optimise the verification assessment seemed like a promising direction to fulfill the requirements of soundness and time-efficiency. We converge towards the sequential use of the following techniques:\\

\noindent {\bf A- Empirical approaches:} \\

To minimize the need for time-consuming formal approaches, the first verification step relies on generating adversarial attacks. If an attack is generated and successful on at least one of the $81$ outputs, the stability of the test point is proven wrong. Attacks intent to `push' the model predictions beyond the bow tie bounds while ensuring that the added perturbations stays within the allowed input noise defined in \S\ref{sec:property}. They are performed for each output $i$ separately i.e.,~we intend to increase $|f_i(x) - f_i(x')|$, independently of the effect on other output indexes. The attack implementation is designed to create a property violation, either a positive (beyond $\overline{\delta_i}$) or a negative one (beyond $\underline{\delta_i}$) . 

We conduct experiments generating several types of classical adversarial attacks, e.g.~Projected Gradient Descent approach (PGD) \cite{madry2017towards} or Fast Gradient Sign Method (FGSM) \cite{goodfellow2014explaining} specifically designed to find local violation of model stability. We make use of the \rm{cleverhans} library\footnote{\url{https://github.com/cleverhans-lab/cleverhans}} and adapt the adversarial loss function to the given local stability property. The goal of this step A is to quickly find examples of model local instability. We notice that a simple attack generation technique such as PGD is sufficient to determine all test points whose local stability property is violated. We do not discard that more advanced attack generation techniques might be required as the models become more locally stable (e.g.~via local stability training). The stability of test points for which no attack was generated is assessed using formal methods.\\

\noindent {\bf B- Incomplete formal methods:} \\

We leverage the less computationally-demanding nature of incomplete methods in order to assess test points whose stability might be less challenging to guarantee. We intend to provide lower and upper bound estimates of the model outputs to perturbed inputs. To do so, we make use of the linear relaxation-based perturbation analysis (LiRPA) verification bound method CROWN \cite{zhang2018efficient}. 

CROWN (or its equivalent DeepPoly \cite{DBLP:journals/pacmpl/SinghGPV19}) is a commonly adopted bound propagation method. It has been shown to provide tighter bounds compared to previously developed techniques by means of linear and quadratic functions which enhance its effectiveness on many activation functions including ReLUs (used in our models). We make use of the version of CROWN implemented within the Airbus open-source decomon\footnote{\url{https://github.com/airbus/decomon}} library \cite{decomon}.  \\

\noindent {\bf C- Complete formal methods:} \\

To evaluate test points whose stability is neither refuted nor guaranteed by the previous methods, we use an in-house Mixed-Integer Linear Program (MILP)-based verifier. Encoding of the network into variables and constraints is done similarly to the Venus library \cite{Botoeva_Kouvaros_Kronqvist_Lomuscio_Misener_2020}. 

The MILP encoding makes use of bounds computed by symbolic interval propagation \cite{wang2018symbinterval}. For each layer $i$ and neuron $j$ in the layer $i$, two continuous variables are created : $y_{i,j,-},y_{i,j,+}$ which corresponds to pre-activation and post-activation values of the neurons. Bounds of these variables are known thanks to the bound propagation function. Layer $i=0$ corresponds to input values. The MILP is encoded as follow: 
\begin{enumerate}
    \item $\forall i\in [1,n_{layers}], j\in [0,n_i], y_{i,j,-}=W_i[j,:] \cdot y_{i-1,:,+}+b_{i,j}$, where $W_i$ and $b_i$ are respectively the model weights and bias of layer $i$
    \item $\forall i\in [1,n_{layers}]$, if layer $i$ is activated by a {\it ReLU}, $y_{i,j,+}=max(y_{i,j,-},0)$ else $y_{i,j,+}=y_{i,j-}$
\end{enumerate}

The $max$ constraint is native into the Gurobi library and can also be chosen instead of the classically used Big-M constraint \cite{Botoeva_Kouvaros_Kronqvist_Lomuscio_Misener_2020}.\\


The negation of the stability property is encoded as follows. Given $x, f(x), \underline{\delta}, \overline{\delta}$ defined in property~\ref{property:stability}:

\begin{enumerate}
    \item Let $\forall j \in [1, 81], indic_{-,j}, indic_{+,j}$ be binary variables. 
    \item $\forall j, indic_{-,j}\rightarrow y_{n_{layers},j,+}\leq f_j(x)+\underline{\delta}_j$
    \item $\forall j, indic_{+,j}\rightarrow y_{n_{layers},j,+}\geq f_j(x)+\overline{\delta}_j$
    \item $\sum_j{(indic_{+,j}+indic_{-,j})}\geq 1$
\end{enumerate}

Each indicator variable will encode the fact that, for one output index, the perturbed NN prediction goes out of the bow-tie property. The last constraint thus encodes that the solver has to find a counterexample violating the property. If the MILP solver finds a solution, we conclude that the property is `False'. If the solver shows the absence of counterexample, we conclude that the property is `True'.\\

The sequence of these verification techniques is illustrated in Fig.~\ref{fig:verif}. While we evaluate the efficiency of this verification process via this specific set of techniques (see \S\ref{sec:experiments}), we expect that similar analysis can be conducted by removing bricks, sequentially combining techniques in another order (e.g.~B+A+C instead of A+B+C) or replacing each brick by any technique falling into the same family (incomplete by incomplete etc.).

In order for the community to be able to perform similar verification technique combination and benchmarking, Airbus has made the verification pipeline open-source. Link and details on the source code are provided in Appendix~\ref{git}.

\section{Experiments}
\label{sec:experiments}

The assessment of local stability (\S\ref{usecase}) is performed on $1000$ test points\footnote{We acknowledge that the present techniques allow the verification of the local stability property of a finite number of points in the input domain and do not permit in their current implementation to prove the global robustness of a continuous input domain. Work is currently being pushed by the formal method community towards exploring the extension of these techniques to global robustness verification.} whose distribution matches the one of the training set and is representative of the operating domain.  \\

The stability evaluation of the $5$ models and $1000$ points are conducted on a machine equipped with an Apple M2 Pro processor and $32$~GB of memory. The MILP-based solver is running with Gurobi 10.0 version. \\

\begin{figure*}[h!]
\centering
\includegraphics[width=12.5cm]{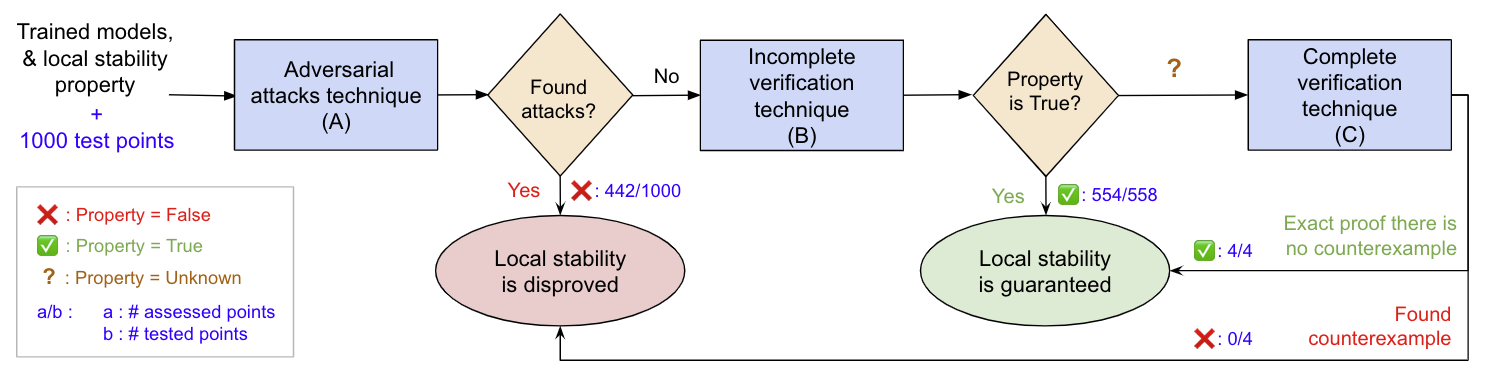}
\caption{Verification pipeline for NN stability assessment.}
\label{fig:verif}
\end{figure*}

Fig~\ref{fig:verif} illustrates the sequence of techniques (empirical, incomplete and complete) used to assess the NN local stability. It also shows how the NN verification unfolds for the model which was trained for $2$ epochs (`2-epoch') i.e.,~it shows the subsets of the $1000$ test points progressing through the different stages (A+B+C), indicating whether their stability property is True (\checkmark), False (X), or unknown (?). We see in this example that adversarial attack generation (A) is successful for $442/1000$ test points. The stability of $554$ of the remaining $558$ test points is then proven via incomplete verification (B). The stability of the last $4$ points is guaranteed via complete verification (C). 

Attacks are generated using the PGD technique using $20$ iterations of step size $\epsilon=0.01$. The MILP verifier is run with a timeout $t = 300$s per point. None of the test points' verification exceeded this timeout.

Table \ref{table-results} summarises the stability assessment performed on two of the models (trained with $2$ and $500$ epochs) using the verification pipeline. As stated earlier, adversarial attack generation can only provide `False' status, but cannot guarantee the property is `True'. Incomplete/CROWN, on the opposite, can only prove that the property is `True' for some test points i.e.,~they guarantee stability but not its lack of. Only complete techniques can unequivocally prove that a point stability is either guaranteed or disproved. 

\begin{table}[h]
\centering
\begin{tabular}{|>{
\hspace{0pt}}m{0.05\linewidth}|>{\hspace{0pt}}m{0.139\linewidth}|>{\hspace{0pt}}m{0.071\linewidth}|>{\hspace{0pt}}m{0.071\linewidth}|>{\hspace{0pt}}m{0.071\linewidth}|>{\hspace{0pt}}m{0.11\linewidth}|>{\hspace{0pt}}m{0.11\linewidth}|>
{\hspace{0pt}}m{0.3\linewidth}|}
\cline{3-8}
\multicolumn{1}{>{\hspace{0pt}}m{0.05\linewidth}}{}& & (1) & (2) & (3) & (4) & (5) & (6) \\
\multicolumn{1}{c}{} & & A & B & C & A+C & B+C&Pipeline \small{A+B+C}\\ 
\hline
\multirow{3}{0.05\linewidth}{\rotatebox[origin=c]{90}{model 2}} 
& \#Tested & 1000 & 1000 & 1000 & 1000/558 & 1000/446 & 1000/558/4 \\ \cline{2-8}
& \#True & -   & 554   & 558   & 558   & 558  & -/554/4 = 558  \\ \cline{2-8}
& \#False    & 442   & -   & 442  & 442  & 442  & 442/-/0 = 442 \\ 
\cline{2-8}
& {\bf Runtime} & {\bf 10.7}  & {\bf 3.3}  & {\bf 267}  & {\bf 19.8}  & {\bf 267} & {\bf 10.7/1.96/3.91 = 16.6}  \\ 
\hline
\multirow{3}{0.05\linewidth}{\rotatebox[origin=c]{90}{model 500}}   
& \#Tested & 1000 & 1000 & 1000 & 1000/552 & 1000/471 & 1000/552/23 \\ \cline{2-8}
& \#True      & -                                            & 529                     & 552                  & 552        & 552          & -/529/23 = 552  \\ 
\cline{2-8}
& \#False    & 448                                          & -                       & 448                  & 448        & 448          & 448/-/0 = 448  \\ 
\cline{2-8}
& {\bf Runtime} & {\bf 11.5} & {\bf 3.4} & {\bf 1091} & {\bf 580} & {\bf 827}          & {\bf 11.5/1.9/307 = 320}\\ 
\hline 
\end{tabular}
\vspace{3mm}
\caption{Stability assessment of $2$ and $500$-epoch models. Columns 1-3 correspond to experiments with only A- adversarial, B- incomplete and C- complete techniques. Columns 4-6 show combinations of these. `\#Tested' refers to the number of data entering a verification brick. `\#True' refers to number of points whose property is guaranteed and `\#False' to the fact that a counterexample is found i.e.,~the NN is not stable. `-' means the technique can not prove the property status. `0' shows it can but did not. Runtimes are provided in seconds.}
\label{table-results}
\vspace{-6mm}
\end{table}

We see that we can generate attacks for $442$ ($448$) out of $1000$ test points for the $2$-epoch ($500$-epoch) model. We can provide local stability guarantees for $554$ ($529$) of the remaining $558$ ($552$) test points. When combining bricks A+B+C, we only have to perform complete verification on the last $4$ ($23$) test points.

\section{Results}
\label{sec:results}

\subsection{A significant verification time gain} 

The stability assessment on the $1000$ test dataset was $3$ to $16$ times faster by combining verification methods instead of solely using complete approaches. The use of MILP-only verification takes $267$s (resp. $1091$s) for the $2$- (resp. $500$-)epoch model. In contrast, the cumulative time for the full pipeline is only $16.6$s ($320$s). 

Similarly, the three-stage pipeline has a significantly faster runtime than a portion of it (e.g.~Bricks A+C) which further shows the effectiveness of combining techniques, thereby reducing the call to time-demanding exact computations. The use of incomplete methods as Brick B allows, e.g.~to converge towards a guarantee of local stability for the large majority of the test data that were still to be evaluated after the adversarial attack stage (e.g.~$554/558$ and $529/552$ for $2$-epoch and $500$-epoch models respectively).

\begin{figure}
\centering
\includegraphics[width=1\textwidth]{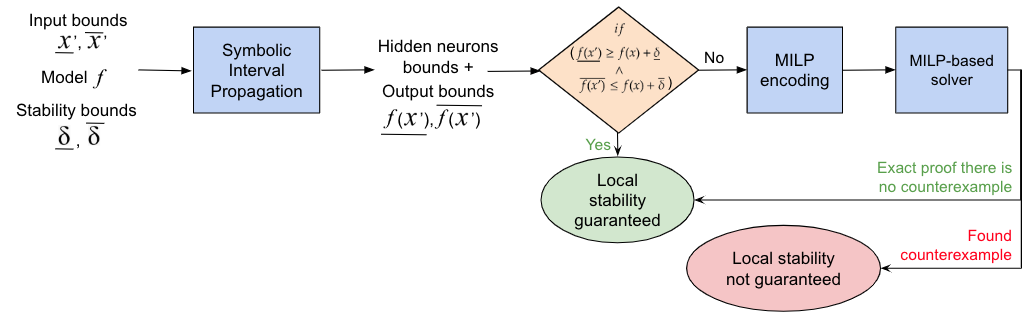}
\caption{Sub-pipeline included in the MILP-based approach}
\label{fig:milppipe}
\vspace{-5mm}
\end{figure}

A subtlety of the complete MILP-based method implemented in the present analysis is that it uses bound propagation \cite{wang2018symbinterval} in order to provide lower and upper values to neurons in the network. When the bound propagation already allow to reach a definite conclusion on the stability property status, there is no call to the MILP solver (see Fig.~\ref{fig:milppipe}). We acknowledge that that the derivation of these bounds are conceptually similar to bounds derived using incomplete methods. While this part of brick C is here performed using symbolic interval analysis that provides tight but potentially looser bounds than CROWD, we are exploring the feasibility of using bounds derived by method B as inputs to C.

An example can be seen in the time of method A+C for the 2-epoch model competitive with respect to the full pipeline (Table~\ref{table-results}, column $4$ vs. column $6$). On this lightly trained model, the first step of bound propagation estimation already concludes on the stability of all but four test points, without the need for additional MILP computations. On the contrary, when using method B+C (column 5), most of the points entering the complete method are not locally stable. They would have been identified during the adversarial attack phase (not used in B+C) and the MILP-based solver has to be called to find or not counter-examples to the property.

We are currently working on making the models more locally stable e.g.~by using stability constrains as part of the training objectives. As the model becomes more and more robust, it will become more challenging to find counterexamples and runtime will increase with the use of formal techniques.

\vspace{-2mm}
\subsection{Insights into the models} 

\begin{figure}[h]
\vspace{-8mm}
\centering
\includegraphics[scale=0.4]{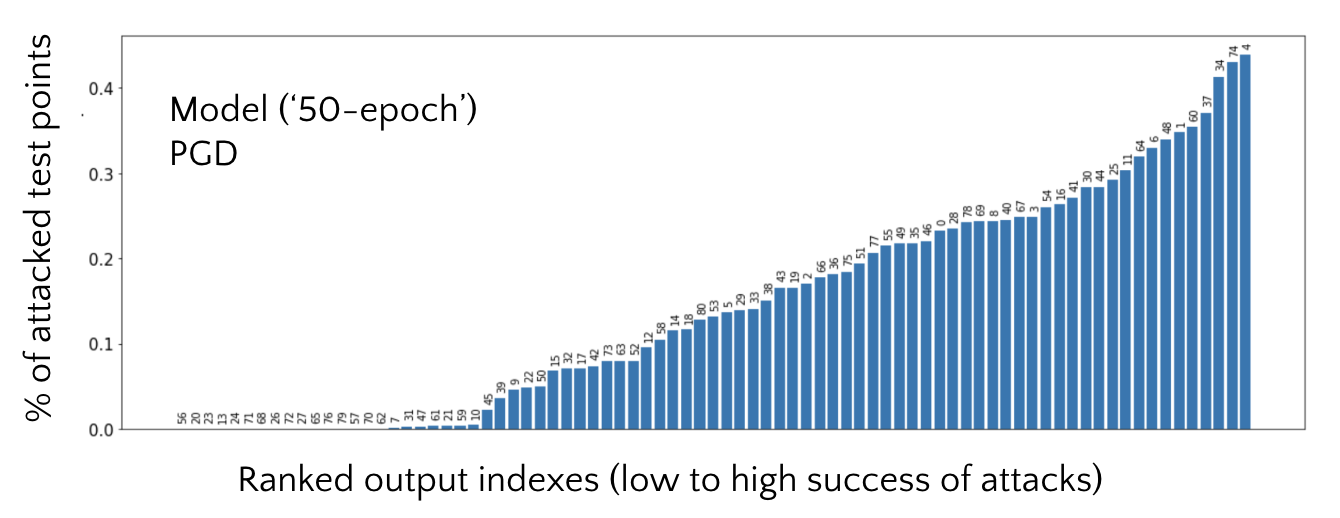}
\caption{Percentage of successful PGD attacks for a given targeted output index (from $0$ to $80$; ranked with increasing success).}
\label{fig:adv}
\end{figure}

Up to $45$\% of the test data could be attacked (with some model-to-model variation). Fig.~\ref{fig:adv} shows the percentage of successful PGD attacks for a given output index. Results are shown for the $50$-epoch model. Attacks are generated for the $1000$ test points and each output index was attacked twice per test point (towards a positive or negative violation) resulting in a total of $2000$ attack attempts.

\begin{figure}
\vspace{-9mm}
\centering
\includegraphics[scale=0.35]{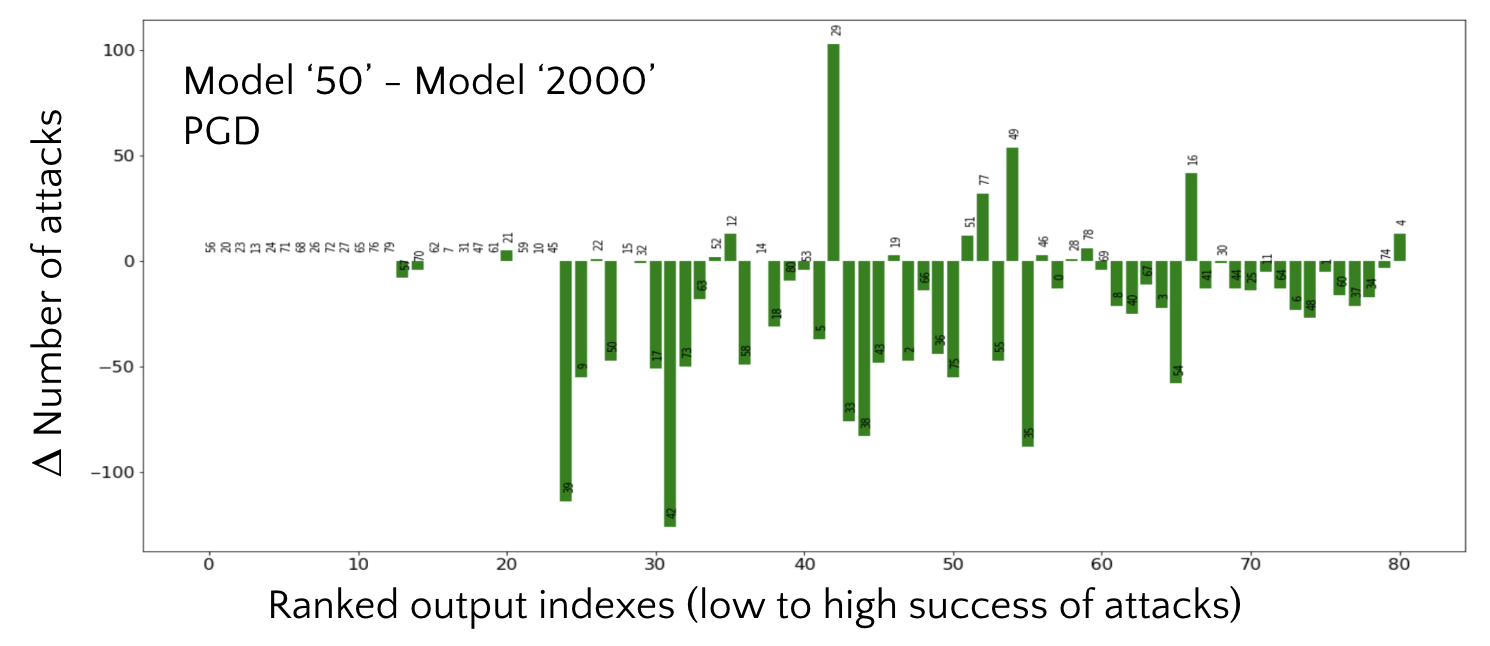}
\caption{Difference of number of attacks (out of $2000$ attempts; see Fig.\ref{fig:adv}) between the $50$-epoch and $2000$-epoch models.}
\label{fig:adv_epochs}
\vspace{-6mm}
\end{figure}

We first observe that some output indexes are more vulnerable than others, a result that is being used for the on-going model strengthening. Indeed, if the local stability is disproved for at least one output index, the test point is considered as not meeting the safety requirement. If some indexes are systematically shown to be vulnerable, the pertinence of these indexes might be re-evaluated.

In this analysis, the adversarial attacks are performed in order of output indexes. We can also greatly benefit from the fact that we can anticipate which indexes are more easily attackable than others to optimise the attack generation on subsequent generation of models and gain time on future robustness assessment.

We observe that the increase of training epochs does not make the models more stable. Table \ref{table-results} shows that the predictions of the $500$-epoch model can be attacked $44.8$\% of the time versus $44.2$\% for the $2$-epoch model. A similar conclusion is reached from Fig.~\ref{fig:adv_epochs} which shows the difference of attacks generated for the $50$ and $2000$-epoch models. We observe however that the stability property is very dependent on the targeted output index. This shows that, in order to build models that are inherently stable (or in general robust to local perturbations), robustness should be part of the training objective and while training with more epochs surely increases performance (up to the point when the model starts overfitting), it does unfortunately not prevent local vulnerabilities.

It is also worth noting that the number of epochs negatively affects the performance of incomplete methods, leading to a longer verification time. 

The experiments also show that counterexamples are preferentially found for $|f_i(x)| > 10$. The requirement in the `knot' stress regime is conservative enough that the property is almost never infringed in the small loads regime.

\vspace{-2mm}
\section{Conclusions and future work}
\label{sec:conclusions}

This case study analysis presents the implementation of a sound multi-technique pipeline for the a posteriori assessment of the local stability property of surrogate NN. By combining the strengths and mitigating the limitations of a number of techniques including adversarial attacks and complete and incomplete formal verification methods, we have not only managed to successfully perform a complete NN stability assessment of the models over the test dataset but also reduced the runtime of the required experiments to do so. Furthermore, the experiments have also shone light on model vulnerabilities and the necessity to investigate the feasibility of looser constraints for the local stability property itself. 

This case study and technique configuration only begins to explore what such a verification pipeline can offer; further tuning of the techniques could indeed lead to an even more time-efficient verification. We performed a number of optimisation experiments e.g.,~hyperparameters tuning for adversarial attacks, MILP-computation etc. But further optimisation work could be done in that respect. As far as the implementation of the MILP solver is concerned (see \S\ref{methodcomb}), the present analysis uses symbolic interval propagation in order to provide starting bounds and restrict the input domain to be verified. For more complex NN however, interval arithmetic could provide excessively conservative bounds that may slow the verification. We will be exploring the use of bounds derived from an incomplete verification tool (e.g.~CROWN) in order to feed them as initial bounds to the complete verifier.

The next step is to enhance NN surrogate models through the integration of empirical and formal components as part of the model design phase (e.g.~during training), moving beyond the sole a posteriori evaluation of its local stability.\\

\bibliographystyle{abbrvnat}
\begingroup
\let\clearpage\relax
\renewcommand\bibpreamble{\vspace{-3\baselineskip}}
\bibliography{main}
\endgroup

\newpage

{\noindent\huge{Appendix}}

\renewcommand\thesection{\Alph{section}}
\renewcommand\thesubsection{\thesection.\arabic{subsection}}
\setcounter{section}{0}

\section{Open-source verification pipeline}
\label{git}

The open-source code of the verification pipeline (AIROBAS) is made available on \href{https://github.com/airbus/Airobas}{github}. \\

\noindent It provides functionalities to:

\begin{itemize}

\item load input data and models,
\item define a robustness property to be verified. The code has been implemented to be as generic as possible and highly parametrizable regarding property definition,
\item design customized verification pipeline, 
\item run verification with a default set of empirical and formal methods e.g.,~attacks using the \rm{cleverhans} library, incomplete verification using LiRPA-based functionalities implemented in \rm{decomon} etc., 
\item append new verification functions or link the code to open-source libraries of one's choice.

\end{itemize}

It also provides a set of step-by-step notebooks to train a number of NN surrogate models (e.g., surrogate to the Rosenbrock analytic function, a runway braking distance estimate etc.), define robustness properties to be assessed and proceed to prove or disprove such properties for a set of test data.

\end{document}